\documentclass{article}
\usepackage[utf8]{inputenc}
\usepackage[T1]{fontenc}
\usepackage[a4paper, total={6in, 8in}]{geometry}
\usepackage{amsmath, amssymb, amsthm}
\usepackage{bm}
\usepackage{graphicx}
\usepackage{xcolor}
\usepackage{textcomp}
\usepackage{natbib}
\usepackage{hyperref}
\usepackage{booktabs}
\definecolor{darkblue}{rgb}{0, 0, 0.5}
\hypersetup{colorlinks=true, citecolor=darkblue, linkcolor=darkblue, urlcolor=darkblue}

\title{\bf A Survey on Neural Abstractive Summarization Methods and Factual Consistency of Summarization Systems}
\author{Meng Cao}
\date{
    School of Computer Science\\
    McGill University\\[2ex]
    November 2021
}

\begin{document}

\maketitle

\section{Introduction}
Automatic summarization is the process of shortening a set of textual data computationally, to create a subset (a summary) that represents the most important pieces of information in the original text. Existing summarization methods can be roughly divided into two types: \textit{extractive} and \textit{abstractive}.
An extractive summarizer explicitly selects text snippets (words, phrases, sentences, etc.) from the source document, while an abstractive summarizer generates novel text snippets to convey the most salient concepts prevalent in the source.

The purpose of this review is to provide a thorough survey of state-of-the-art abstractive summarization approaches and discuss some of the challenges these methods face.
We focus on the abstractive summarization task because it is computationally more challenging than extraction methods and is closer to the way humans write summaries.
There are two parts in this survey. In the first part, we will explore some classic as well as recent summarization methods. The focus will be on neural network-based abstractive summarization methods.
Firstly, we will briefly review some non-neural abstractive summarization methods from the pre-neural network era. Most of these methods use either deletion-based \citep{knight2002summarization} or statistical models \citep{10.3115/1075218.1075259}.
Secondly, we examine in detail five neural-based abstractive text summarization models \citep{rush-etal-2015-neural, chopra-etal-2016-abstractive, nallapati-etal-2016-abstractive, see-etal-2017-get, gehrmann-etal-2018-bottom, paulus2018a}.
These methods all adapt recurrent neural network (RNN) or convolutional neural network (CNN) architecture for sentence representation.
With the help of the representational capabilities of neural networks and large training data sets, these models are largely superior to previous non-neural summarization methods. 
Then, we will move on to the most recent work on abstractive summarization using large pre-trained language models \citep{liu-lapata-2019-text, lewis-etal-2020-bart, pmlr-v119-zhang20ae, stiennon2020learning}. Different from previous RNN-based methods, these models are based on the Transformer architecture \citep{NIPS2017_3f5ee243} and pre-trained on a large amount of raw text before being fine-tuned on the summarization task.

In the second part, we will discuss some recent work on the factual consistency evaluation of abstractive summarization systems.
The factual consistency of a summary is determined by its agreement with the facts in the input document.
For the extractive system, its factual consistency is mostly ensured since all sentences are extracted from the source. For abstractive summarization models, however, they are prone to generate statistically likely but actually inconsistent summaries. Common factual errors include manipulating information in the source document and adding information that cannot be directly inferred from the source. 
Factual errors are especially common in large-scale Transformer-based abstractive summarization models pre-trained on a large amount of online texts.


\section{Background}
\subsection{The Automatic Summarization Task}
Summarization systems are designed to take a single document, a cluster of news articles, a dialogue, or an
email thread as input, and produce a concise summary of the most important information in the origin content \citep{nenkova2011automatic}.

Most summarization systems can be roughly divided into two categories: \textit{extractive} and \textit{abstractive}. Extractive summarization approach generates a summary by selecting important segments from the original document and putting them together to form a coherent summary. Abstractive methods first build an internal semantic representation of the original text, and then use this representation to create a condensed summary that captures the core information of the text. Compared with the extractive summarization approach, the abstractive summarization method can generate more concise summaries. For example, an summary that is shorter than any sentence in the source document. However, abstractive summarization is computationally more challenging than extractive summarization and requires a deep understanding of the original content.

\subsection{Datasets}
In this section, we will discuss four commonly used summarization datasets. These datasets are all in English and are used for the single-document summarization task. Note that there are many other different summarization datasets that are designed for different tasks (e.g. medical summarization, web summarization). We choose these four because they are more representative.

\paragraph{DUC-2004} This dataset is introduced in the DUC-2004\footnote{\url{https://duc.nist.gov/duc2004/}} summarization competition for evaluation. The dataset consists of 500 news articles from the New York Times and Associated Press Wire services. Each article is paired with 4 different human-written reference summaries.
DUC-2004 is designed to focus on very short ($\leq$ 75 bytes) summaries, thus incentivizing participants to go beyond extraction methods.

\paragraph{Gigaword \citep{graff2003english, napoles-etal-2012-annotated}}
Gigaword is a headline generation corpus consisting of around 4 million title-article pairs. The articles and summaries in Gigaword are shorter than those in DUC-2004.
The news articles in the dataset are sourced from various domestic and international news services. 

\paragraph{CNN/DailyMail \citep{nallapati-etal-2016-abstractive}} The CNN/DailyMail dataset contains over 300k unique news articles as written by journalists at CNN and the DailyMail. The original dataset is created for machine reading and comprehension and abstractive question answering. \cite{nallapati-etal-2016-abstractive} concatenate all the summary bullets of each article in
the original order to obtain a multi-sentence summary. CNN/DailyMail can be used to train models for abstractive and extractive summarization. However, the dataset is more extractive in nature.

\paragraph{\textsc{XSum} \citep{narayan-etal-2018-dont}} In order to create a large-scale abstractive summarization dataset that does not favor extractive models, \cite{narayan-etal-2018-dont} collect 226,711 online articles from the British Broadcasting Corporation (BBC). Each BBC article starts with an one-sentence summary, which can be used as an short news summary that captures the most important information of the article. Unlike the headlines in Gigaword, the purpose of which is to encourage readers to read the story; the summaries in \textsc{XSum} utilize information scattered in various parts of the document.
The scale and the abstractiveness of the \textsc{XSum} dataset make it suitable for training and evaluating abstractive summarization systems. However, it is pointed out that some summaries in \textsc{XSum} contain information that cannot be inferred from the source document.

\subsection{Evaluation of Summarization Systems}
They are the two main ways for evaluating abstractive summarization systems: one is to collect human judgments on the quality of the output summary, and the other is to compare the output summary with a human-written gold standard. Human judgement is arguably the best overall but it is time-consuming and expensive. In this section, we will discuss the most commonly used automatic measures for summarization system evaluation: ROUGE scores.

\subsubsection[\textsc{Rouge}]{\textsc{Rouge} \citep{lin-2004-rouge}}
\textsc{Rouge} stands for Recall-Oriented Understudy for Gisting Evaluation. It is one of the most commonly used automatic evaluation metric for summarization systems. \textsc{Rouge} calculates the degree of overlap between the generated summary and the reference summary. There are a few variants of \textsc{Rouge} scores.

\textsc{Rouge-N} calculates the percentage of overlapped $n$-gram between the generated summary and the gold standard reference summary. It is defined as follows:
\begin{equation*}
    \textsc{Rouge-N} = \frac{\sum_{s \in S_{\text{reference}}} \sum_{\text{gram}_n \in s} \text{Count}_{\text{match}}(\text{gram}_n)}{\sum_{s \in S_{\text{reference}}} \sum_{\text{gram}_n \in s} \text{Count}(\text{gram}_n)}
\end{equation*}
where $S_{\text{reference}}$ is the reference summary set, $n$ stands for the length of the $n$-gram, $\text{gram}_n$, $\text{Count}$ calculates the number of times $\text{gram}_n$ appears in the reference summary, $\text{Count}_{\text{match}}$ is the number of $n$-grams co-occurrences in the output summary
and the reference summary.
In summary, the numerator counts the number of overlapping $n$-grams found in both the output summary and the reference. The denominator counts the total number of $n$-grams in the reference.
This is a recall-related measure to ensure that the output summary captures all the information contained in the reference. To punish the model for generating very long summaries, a precision-related metric is calculated by replacing the denominator with the total number of $n$-grams in the output summary. With both precision and recall scores, we can calculate the more balanced \textsc{Rouge} F1 measure: $$\textsc{Rouge}_{\text{F1}} = 2\frac{\textsc{Rouge}_{\text{P}} * \textsc{Rouge}_{\text{R}}}{\textsc{Rouge}_{\text{P}} + \textsc{Rouge}_{\text{R}}}$$
\textsc{Rouge-L} measures the longest common subsequence (LCS) between the output summary and the reference summary. The idea is that a longer shared sequence indicates a higher similarity between the two summaries. \textsc{Rouge-L} is calculated in almost the exact same way as \textsc{Rouge-N}, but replace the common $n$-gram counts in the numerator with the length of the longest common subsequence.



\section{Abstractive Summarization Techniques}
We roughly divide the existing summarization methods into two categories according to whether neural networks are used. For neural network-based models, we further classified them based on whether they have been pre-trained on raw text in a semi-supervised manner. Since almost all pre-trained models use the Transformer architecture, we refer to pre-trained summarization models as the Transformer-based models. In Section \ref{sec:rl}, we will talk about RL-based summarization methods.


\subsection{Pre-Neural Network Era}
\subsubsection[Statistical Approaches]{Statistical Approaches \citep{10.3115/1075218.1075259}}
One way to look at abstractive summarization is to think of it as a problem analogous to statistical machine translation. Given an input document $x$, the summarization task is to find:
\begin{equation*}
    y^* = \arg \max_{y} p(y \mid x; \theta)
\end{equation*}
where $\theta$ is the parameter of the statistical model. The most important question is how to model $p$. \cite{10.3115/1075218.1075259} 's approach consists of two major steps: \textit{content selection} and \textit{surface realization}. The task of content selection is to estimate the likelihood of some tokens appearing in the summary given the source document. This probability can be estimated as the product of the likelihood of (i) the token being selected for the summary, (ii) the length of the summary, and (iii) the most likely sequencing of the selected tokens:
\begin{equation*}
\begin{split}
    p(w_1, \dots, w_n) = & \prod_{i=1}^n p(w_i \in y \mid w_i \in x) \\
    & \cdot p(\text{len}(x) = n) \\
    & \cdot \prod_{i=2}^n p(w_i \mid w_1, \dots, w_{i-1}) \\
\end{split}
\end{equation*}
where $p(\text{len}(x) = n)$ can be modeled using a Gaussian distribution and $ p(w_i \in y \mid w_i \in x) $ is computed using the Bayes' Rule:
\begin{equation*}
    p(w_i \in y \mid w_i \in x) = \frac{p(w_i \in x \mid w_i \in y) \cdot p(w_i \in y)}{p(w_i \in x)}
\end{equation*}
For surface realization, the probability of any particular surface ordering as a summary can be estimated using a bi-gram language model. Combining the content and summary structure generation, the overall scoring function is as follows:
\begin{equation*}
\begin{split}
    \arg \max_{y} ( & \alpha \cdot \sum_{i=1}^n \log{(p(w_i \in y \mid w_i \in x))} \; + \\
    & \beta \cdot \log{(p(\text{len}(y) = n))} \; + \\
    & \gamma \cdot \sum_{i=2}^n \log{(p(w_i\mid w_{i-1}))})
\end{split}
\end{equation*}
where $\alpha, \beta$ and $\gamma$ are hyper-parameters learned through cross-validation. 

In this work, the content selection and word ordering are jointly applied using a statistical model, which is learned from a training corpus.
The neural network-based summarization models that appeared later used similar ideas. 

\subsubsection[Deletion-Based Methods]{Deletion-Based Methods \citep{knight2002summarization}}
\cite{knight2002summarization} focus on a scaled-down version of the text summarization problem: \textit{sentence
compression}. The task setup is as follows: take a natural language sentence as input, delete any subset of words in the input sentence and ensure the remaining words (order unchanged) form a grammatical compression. They first convert the input sentence into a grammar tree and perform the rewriting operation by applying a series of shift-reduce-drop operations on the tree. One limitation of this deletion-based method is the inability to generate new words and novel sentence structures.

\subsection{CNN-/RNN-based Methods}
\label{sec:rnn_based_methods}

\subsubsection[ABS]{ABS \citep{rush-etal-2015-neural}}
The distribution of interest, $p(y_{i+1}|x, y_c; \theta)$, is
a conditional language model based on the input text $x$ and previously generated tokens $y_c$. The core of \citep{rush-etal-2015-neural}'s work is to directly parameterize the distribution as a neural network. This opens a door for neural network-based summarization systems. Specifically, their network contains a neural language model for estimating the contextual probability of the next token
and an encoder module which acts as a conditional summarization model.

The authors explore three types of encoder models: \textit{Bag-of-Words Encoder}, \textit{Convolutional Encoder} and \textit{Attention-Based Encoder}. We will discuss each type of encoder model in detail.
\paragraph{Bag-of-Words Encoder}
The bag-of-words encoder simply calculate the average of input token embeddings without considering properties of the word order or relationships: $f_{\textrm{enc}}(x, y_c) = \frac{1}{M} \sum_{i=0}^M {F}x_i = \frac{1}{M} \sum_{i=0}^M {\tilde{x}}_i$,
where ${F} \in \mathbb{R}^{H \times V}$ is the input-side word embedding matrix and $x_i \in \{0,1\}^{V}$ is an one-hot vector with size $V$. Note that ${F}$ is the only parameter of the bag-of-words encoder.
\paragraph{Convolutional Encoder} This encoder uses deep convolutional network to encode the input sentence. The convolutional encoder has $L$ layers in total. At each layer, the context tokens are first encoded with a 1D convolution layer: $\Bar{x}_i^l = {Q}^l {\tilde{x}}^{l-1}_{[i-Q, \dots, i+Q]}$, where ${\tilde{x}}^{l-1}$ is the previous layer's output (or the word embeddings for the first layer); $l$ is the layer number and ${Q}$ is the context window size. Then, there is a 2-element temporal max pooling layer and a pointwise non-linearity layer: $\tanh(\max \{\Bar{x}^l_{2i-1}, \Bar{x}^l_{2i}\})$. On the top, a max-pooling-over-time is performed to obtain the final input text representation.

\paragraph{Attention-Based Encoder}
Inspired by the attention mechanism of \citep{DBLP:journals/corr/BahdanauCB14} in machine translation, the authors also apply a attention-based encoder to learn a latent soft alignment between the input text and the generated summary. The model can be written as follows:
\begin{equation*}
\begin{split}
    f_{\textrm{enc}}(x, y_c) &= {p}^T {\Bar{x}}, \\
    {p} & \propto \exp{({\tilde{x}} {P} {\tilde{y}}_c^t)},\\
    \forall{i}, \;\; {\Bar{x}}_i &= \sum_{j=i-Q}^{i+Q} {\tilde{x}}_j / Q.
\end{split}
\end{equation*}
where ${\tilde{x}}$ is the input-side word embeddings and  ${\tilde{y}}_c^t$ is the output-side embeddings with a context size equals to $C$: ${\tilde{y}}_c^t = [{y}_{t-C+1},\dots,{y}_t]$.
Informally, we can think of this model as replacing the uniform distribution in the bag-of-words encoder with a learned alignment distribution ${p}$.

\paragraph{Decoder} For estimating the contextual probability of
the next token, \citet{rush-etal-2015-neural} adapts a feed-forward neural network language model (NNLM). The full model can be described as:
\begin{equation*}
\begin{split}
p({y}_{i+1}\mid {y}_c, {x}; \theta) & \propto \exp ({V}{h} + {W} f_{\textrm{enc}}({x}, {y}_c)), \\
h &= \tanh({U}{\tilde{y}}_c).
\end{split}
\end{equation*}
where ${U}, {V}, {W}$ are model parameters.

\paragraph{Training} In \cite{rush-etal-2015-neural}, the authors train the summarization model by minimizing the negative log-likehood loss using mini-batch stochastic gradient descent.
\begin{equation*}
\begin{split}
    \text{NLL}(\theta) &= -\sum_{j=1}^J \log{p({y}^{(j)} \mid {x}^{(j)}; \theta)} \\
    &=-\sum_{j=1}^J \sum_{t=1}^{N-1} \log{p({y}^{(j)}_{t+1} \mid {x}^{(j)}, {y}_c; \theta)} \\
\end{split}
\end{equation*}
where ${y}_c$ is the gold standard contexts (teacher forcing). $J$ is the size of the mini-batch. All other work in this section adopts the same training method.

One limitation of this work is that the encoder is limited by the context window.
Next, we see how \cite{chopra-etal-2016-abstractive} use recurrent neural networks to solve this limitation.

\subsubsection[RAS-LSTM \& RAS-Elman]{RAS-LSTM \& RAS-Elman \citep{chopra-etal-2016-abstractive}}
\citet{chopra-etal-2016-abstractive} improves \citet{rush-etal-2015-neural}'s work by replacing the feed-forward neural network decoder (NNLM) with recurrent neural network (RNN). 
Compared with NNLM, RNN can model arbitrarily long contexts within its capacity, without the need to manually select the context length $C$. The previous context information is stored in RNN's hidden state $h_t$.
The RNN is used to model the following conditional probability distribution $p_t$:
\begin{equation*}
    p(y_{t}\mid \{y_1,\dots,y_{t-1}\}, {x}; \theta) = g_{\theta}(h_t, c_t)
\end{equation*}
where $c_t$ is the encoder output and $h_t$ is the hidden state of the RNN:
\begin{equation*}
    h_t = g_{\theta}(y_{t-1}, h_{t-1}, c_t)
\end{equation*}

The authors explore two types of RNN decoders: \textit{Elman RNN} \citep{elman1990finding} and \textit{LSTM} \cite{HochSchm97}.

\paragraph{Elman RNN} The Elman RNN is defined as:
\begin{equation*}
\begin{split}
    h_t &= \sigma (W_1 y_{t-1} + W_2 h_{t-1} + W_3 c_t) \\
    p_t &= \textrm{softmax} (W_4h_t + W_5c_t)
\end{split}
\end{equation*}
where $W_{\{1,2,3,4,5\}}$ are all learnable parameters of the neural network. 

\paragraph{LSTM} The LSTM decoder is more complicated than Elman RNN with more internal layers:
\begin{equation*}
\begin{split}
    i_t &= \sigma (W_1 y_{t-1} + W_2 h_{t-1} + W_3 c_t) \\
    i'_t &= \tanh (W_4 y_{t-1} + W_5 h_{t-1} + W_6 c_t) \\
    f_t &= \sigma (W_7 y_{t-1} + W_8 h_{t-1} + W_9 c_t) \\
    o_t &= \sigma (W_{10} y_{t-1} + W_{11} h_{t-1} + W_{12} c_t) \\
    m_t &= m_{t-1} \odot f_t + i_t \odot i'_t \\
    h_t &= m_t \odot o_t \\
    p_t &= \textrm{softmax} (W_{13}h_t + W_{14}c_t)
\end{split}
\end{equation*}
where $W_{\{1,2,\dots,12\}}$ are learnable parameters of the decoder. Operation $\odot$ refers to component-wise multiplication.


\subsubsection[Hierarchical Attentive RNNs]{Hierarchical Attentive RNNs \citep{nallapati-etal-2016-abstractive}}
Compared with \cite{rush-etal-2015-neural} and \cite{chopra-etal-2016-abstractive}'s approaches, \cite{nallapati-etal-2016-abstractive}'s work introduces four major changes: first, it applies RNN for both the encoder and decoder network. Second, the input to the encoder model is not only the word-embeddings-based representation but also includes other linguistic features such as part-of-speech tags, named-entity tags, and TF and IDF statistics of the words. Third, the model's decoder adopts the pointer mechanism to deal with rare words. Finally, they design a hierarchical attention structure for the decoder to deal with long source documents. We will discuss each point in detail.
\paragraph{RNN Encoder-Decoder} In \cite{nallapati-etal-2016-abstractive}, the authors adapt a bidirectional GRU-RNN \citep{69e088c8129341ac89810907fe6b1bfe} as the encoder and an uni-directional GRU-RNN with the same hidden layer size as the decoder. They also apply attention mechanism over the hidden states of the encoder at each time step. Another technique they used is called the large vocabulary trick (LVT) \citep{jean-etal-2015-using}. The idea is to limit the decoder vocabulary to only words that appear in the source document in the current mini-batch training. In order to make the vocabulary size fixed, the most commonly used words in the target dictionary are also added. The LVT technique is used to reduce the softmax layer computation and speed up the training.

\paragraph{The Pointer Mechanism} One problem in summarization is that the input document often contains rare words such as key words or named-entities that do not exist in the vocabulary. At that time, the word-piece tokenizer was not widely used. The pointer mechanism is proposed to enable the model to directly copy those out-of-vocabulary words (OOVs) from the input. At each decoding step, the decoder decides to either generate a word from the vocabulary or copy a token from the source document. This is sometimes also referred to as the ``copy mechanism''. Based on the entire available context at each time step, the copying probability is modeled as a sigmoid activation function over a linear layer as follows:
\begin{equation*}
    P(s_i=1) = \sigma ({v} \cdot ({W}_h {h}_i + {W}_y {y}_{t-1} + {W}_c {c}_i + {b}))
\end{equation*}
where ${h}_i$ is the RNN hidden state, ${y}_{t-1}$ is the embedding of previous generated token, ${c}_i$ is the attention-weighted context vector, ${W}_h, {W}_y, {W}_c, {b}$ and ${v}$ are learnable parameters. $s_i$ is an indicator variable that indicates whether to copy a token from the input document. They reuse the attention weight on the document word as the probability of that the word being copied.

\paragraph{Hierarchical Attention} To deal with very long input document, \cite{nallapati-etal-2016-abstractive} proposes a hierarchical attention structure to capture both word-level and sentence-level information. There are two bi-directional RNNs on the source side in their model. One runs on the word-level and the other runs on the sentence-level. The word-level attention is further weighted by the sentence-level attention. Finally, the re-normalized word-level attention is used to compute the attention-weighted context vector ${c}_i$.


\subsubsection[Pointer-Generator Network]{Pointer-Generator Network \citep{see-etal-2017-get}}
\cite{see-etal-2017-get}'s work is similar to \cite{nallapati-etal-2016-abstractive}'s model, they all use a RNN-based sequence-to-sequence architecture, and both use the pointer mechanism to handle OOV words in the source document. There are two main differences in \cite{see-etal-2017-get}'s work. First, the authors mix the probabilities from the copy distribution and the vocabulary distribution when decoding: $P(w) = p_{\textrm{gen}}{P}_{\textrm{vocab}}(w) + (1-p_{\textrm{gen}})\sum_{i:w_i=w}a_i^t$, where $w$ is OOV words in the document and $a_i^t$ is the attention weight at decoding step $t$. 

Second, \cite{see-etal-2017-get} introduces coverage mechanism to tackle the repetition problem in abstractive summarization. They maintain a \textit{coverage vector} $c^t$, which is the sum of attention distributions over all previous decoder time-steps: $c^t = \sum_{t'=0}^{t-1} a^{t'}$. Intuitively, $c^t$ indicates how much attention each word in the source document has received so far. The coverage vector is used as an additional input to the attention mechanism. In order to penalize the model for repeatedly attending to the same positions, they add a new \textit{coverage loss} term:
\begin{equation*}
    \textrm{covloss}_t = \sum_i \min(a_i^t, c_i^t)
\end{equation*}
The experiment results show that coverage loss is effective for eliminating repetition and improving ROUGE scores.


\subsubsection[Bottom-Up Summarization]{Bottom-Up Summarization \citep{gehrmann-etal-2018-bottom}}
\cite{gehrmann-etal-2018-bottom} proposes a \textit{bottom-up} approach for abstractive summarization. They use the same model architecture as \cite{see-etal-2017-get}. Their method consists of two steps: first, they apply a content selection system to decide on relevant parts of the source document. This is framed as a sequence-tagging task with the objective of identifying relevant tokens in the source document. In the second step, they use a mask to limit the copied words to the selected parts of the text. 

\paragraph{Content Selection}
The content selection step is framed as a binary sequence-tagging task. For each token in the source document, it is labeled as either 1 or 0. 1 if the word is relevant and 0 otherwise. To create a training set for this task, the authors align the summaries to the document.
For each word $x_i$ in the document, it is labeled as relevant if (1) it is part of the longest possible sub-sequence of tokens $s$ if $s \in x$ and $y \in y$, and (2) $s$ is the first occurrence in the document.
They train a bidirectional LSTM model for this tagging task.

\paragraph{Bottom-Up Copy Attention}
During training, the pointer-generator model as well as the content selector model are trained on the complete dataset without masking. At inference time, they first apply the trained content selector to predict a probability $q_i$ of each token being selected. Then, this selection probability is used to adjust the attention distribution over source document tokens. Let $a_j^i$ be the attention score at decoding step $j$ for word $i$. The adjusted attention score $\tilde{a}_j^i$ is as follows:
\begin{equation*}
  \tilde{a}_j^i =
    \begin{cases}
      a_j^i & \text{if $q_i > \epsilon$} \\
      0 & \text{otherwise} \\
    \end{cases}
\end{equation*}
where $\epsilon$ is a pre-determined threshold. The authors also re-normalize the adjusted attention scores to make sure its a proper probability distribution.


\subsection{Transformer-based Methods}
In this section, we will discuss three Transformer-based \cite{NIPS2017_3f5ee243} summarization models.
Unlike the methods mentioned in the Section \ref{sec:rnn_based_methods} where the models are trained from scratch on the summarization dataset, the models we will discuss in this section is first pre-trained on a large corpus in an unsupervised manner before fine-tuning on the summarization dataset. The pre-training stage helps the model learn rich syntactic and semantic knowledge, which greatly improves the performance of the model on the summarization task. 

\subsubsection[\textsc{BertSum}]{\textsc{BertSum} \citep{liu-lapata-2019-text}}
Compared with \cite{nallapati-etal-2016-abstractive} and \cite{see-etal-2017-get}'s work, \cite{liu-lapata-2019-text} also adapts a sequence-to-sequence architecture but replace the RNN-based encoder \& decoder with the Transformer architecture. 

\paragraph{Transformer Encoder}
They use a 6-layered Transformer encoder model. All six layers are identical (but different parameters) and each layer consists of two sub-layers, which are multi-head self-attention network and fully connected feed-forward network. Each sub-layer has a residual connection and normalization, so the output of the sub-layer can be expressed as:
\begin{equation*}
    h^{l+1} = \textrm{LayerNorm}(h^l + (\textrm{SubLayer}(h^l)))
\end{equation*}
where $h^l$ is previous sub-layer's output. Next, we will discuss the two sub-layers in order:
\begin{enumerate}
    \item \textit{Multi-head self-attention}: Multi-head attention uses $h$ different linear transformations to project $Q, K$ and $V$, and finally concatenate the outputs together:
    \begin{equation*}
    \begin{split}
        \textrm{MultiHead}(Q,K,V) &= \textrm{Concat}(\textrm{head}_1,\dots,\textrm{head}_h)W^O \\
        \textrm{head}_i &= \textrm{Attention}(QW_i^Q, KW_i^K, VW_i^V)
    \end{split}
    \end{equation*}
    where $h$ is the number of heads and $\{W_i^Q, W_i^K, W_i^V$, $W^O\}$ are attention parameters. The attention function is as follows:
    \begin{equation*}
        \textrm{Attention}(Q, K, V) = \textrm{softmax}(\frac{QK^T}{\sqrt{d_k}})V
    \end{equation*}
    where $d_k$ is the dimension of $K$.
    \item \textit{Position-wise feed-forward networks}: This sub-layer contains a fully connected feed-forward network:
    \begin{equation*}
        \textrm{FNN}(x) = \max (0, xW_1 + b_1)W_2 + b_2
    \end{equation*}
\end{enumerate}

\paragraph{Transformer Decoder} Similar to the encoder, the decoder also consists of a stack of $N = 6$ identical layers. There are three sub-layers in each decoder layer, two of which are the same as the two sub-layers in the encoder. The decoder also inserts a third sub-layer that performs multi-head attention on the output of the encoder stack. In this sub-layer, the $K$ and $V$ are from encoder task output and $Q$ comes from the output of the previous layer of the decoder.

\paragraph{BERT} Bidirectional Encoder Representations from Transformers (BERT; \cite{devlin-etal-2019-bert}) is a Transformer-based language representation model,
trained using the Masked Language Modeling (Masked LM) task and the Next Sentence Prediction (NSP) task. The pre-training corpus consists of BooksCorpus (800M words) and English Wikipedia (2,500M words).
For the Masked LM task, the model is trained to predict 15\% of randomly masked token in the input sentence. The masked tokens are replaced with a \texttt{[MASK]} symbol. For the NSP task, each training example consists of sentences $A$ and $B$. There is a 50\% chance that $B$ will be the actual next sentence follows $A$, and there is a 50\% chance that $B$ is a random sentence sampled from the corpus. The model is trained to predict $B$ follows $A$ or not.

\textsc{BertSum} uses the pre-trained BERT model as the encoder. In order to learn better sentence representation, \textsc{BertSum} inserts multiple \texttt{[CLS]} tokens at the beginning of each document sentence and using interval segmentation embeddings. For decoder, \textsc{BertSum} adopts a 6-layered randomly initialized Transformer decoder model. In addition, \cite{liu-lapata-2019-text} propose a two-stage fine-tuning method: the encoder is first fine-tuned on the extractive summarization task and then on the abstractive summarization task.


\subsubsection[BART]{BART \citep{lewis-etal-2020-bart}}
Unlike BERT, which is a masked language model built using only Transformer encoders, BART is a sequence-to-sequence denoising autoencoder built using Transformer-based encoder and decoder. Because of the sequence-to-sequence nature of BART, it is particularly effective when fine tuned for text generation tasks (e.g. machine translation, text summarization). The pretraining process of BART consists of two steps (1) define an arbitrary noising function to corrupt the original sentence, and (2) train BART to reconstruct the original text. They use standard token-level cross-entropy loss for the reconstruction loss. There are five types of corruption functions introduced in the paper:
\begin{enumerate}
    \item \textit{Token Masking}: Following BERT \cite{devlin-etal-2019-bert}, randomly sample 15\% tokens and mask them with \texttt{[MASK]}.
    \item \textit{Token Deletion}: Random tokens are deleted from the input. No \texttt{[MASK]} tokens are used.
    \item \textit{Text Infilling}: Sample a number of text spans, and the span lengths are drawn from a Poisson distribution ($\lambda=3$). Each sampled span is replaced with a single \texttt{[MASK]} token.
    \item \textit{Sentence Permutation}: Randomly shuffle the order of sentences in the input document.
    \item \textit{Document Rotation}: Randomly select a token in the input document, and rotate the document so that it starts with that token. This task is to predict the  original beginning of the document.
\end{enumerate}

On CNN/DailyMail, BART large model achieves 44.16 ROUGE-1 score. On \textsc{XSum}, the ROUGE-1 score is 45.14. Both results are much higher than RNN-based models. They also compare five pre-training methods in isolation on CNN/DailyMail and \textsc{XSum} datasets. The results indicate that \textit{Text Infilling} is the most effective one among five objectives. Rotating documents or permuting sentences perform poorly in isolation.

\subsubsection[PEGASUS]{PEGASUS \citep{pmlr-v119-zhang20ae}}
PEGASUS is a Transformer-based encoder-decoder model specifically pre-trained for the summarization task. It uses the same model architecture as the BART \cite{lewis-etal-2020-bart} model. \cite{pmlr-v119-zhang20ae} propose a new pre-training task: Gap Sentences Generation (GSG). The idea is to mask whole sentences from documents, and train a language model to generate masked sentences based on the rest parts of the documents.
Each masked sentence is replaced with a special \texttt{[MASK1]} token to inform the model. They propose three strategies to select sentences for masking: (1) \textit{random}: uniformly select $m$ sentences at random. (2) \textit{lead}: select the first $m$ sentences and (3) \textit{principal}: select top-$m$ scored sentences according to the ROUGE F1 score between the sentence and the rest of the document. On \textsc{XSum}, PEGASUS is slightly better than BART in terms of ROUGE scores.


\subsection{Text Summarization with RL}
\label{sec:rl}
In RL, the agent is trained to take actions in an environment in order to maximize the notion of cumulative reward. For the summarization task, reinforcement learning offers more flexibility as we can design different reward functions to focus on different aspects of the model. (e.g. factual consistency).

From RL perspective, the summary generation process can be viewed as a finite Markov Decision Process (MDP). At each time-step $t$, the state $s_t=(y_{<t}, x)$, where $x$ is the input document and $y_{<t}$ are previously generated tokens. The agent, which is the summarization model, takes an action by generating a new token $y_t$. Depending on the action taken, the agent gets a reward $r_t = R(s_t, y_t)$ and deterministically transition to the next state $s_{t+1}=(y_{<t+1}, x)$. The probability of taking each action (i.e. token) is specified by the policy $\pi(y_t \; | \; s_t)$. The goal of the agent is to maximize the cumulative reward throughout the trajectory:
\begin{equation*}
\label{eq:traditional_obj}
    J(\pi) = \mathbb{E}_{\tau \sim \pi} \Big[ \sum_{t=0}^T \gamma^t r_t \Big]
\end{equation*}
where $\gamma$ is the discount factor.
In text generation, the reward is usually obtained when the entire sequence is generated (i.e. $r_{t < T}=0$). In this section, we will talk about two recently proposed RL algorithm for abstractive summarization.

\subsubsection[Deep Reinforced Model]{Deep Reinforced Model \citep{paulus2018a}}
\cite{paulus2018a} proposes a hybrid learning objective function which combines the ``teacher forcing'' objective \citep{6795228} and the self-critical policy gradient training objective \citep{rennie2017self}. In the standard teacher forcing training objective, the model is trained to minimize a negative log-likelihood loss conditioned on ground-truth summary at each decoding step:
\begin{equation*}
    L_{ml} = - \sum_{t=1}^n \log p(y_t^* \mid y_1^*, \dots, y_{t-1}^*, x)
\end{equation*}
where $y^*=\{y_1^*,y_2^*,\dots,y_n^*\}$ is the ground-truth reference sequence for a given input document $x$. There are two main problems with this teacher forcing objective. The first problem is known as the \textit{exposure bias} \citep{DBLP:journals/corr/RanzatoCAZ15}. 
It comes from the fact that the language model is trained to predict the next token given the previous ground-truth tokens as input. However, at inference time, the model does not have this supervision, and the generated tokens must be fed back as input at each time step. This discrepancy makes the model brittle at inference time and errors will quickly accumulate. The second issue is that the teacher forcing objective does not always correlate to the discrete evaluation metric such as ROUGE since the ROUGE metrics do not take token orders into account.

On way to mitigate these problems is to use reinforcement learning to directly optimize the target evaluation metric. Now consider two sequences $y^s$ and $\hat{y}$. $y^s$ is obtained by sampling from the $p(y_t^s \mid y_1^s,\dots,y_{t-1}^s,x)$ probability distribution at each decoding time step and $\hat{y}$ is obtained using greedy decoding: $\arg \max_{\hat{y}_t} p(\hat{y}_t \mid \hat{y}_1,\dots,\hat{y}_{t-1},x)$. The RL objective is given as follows:
\begin{equation*}
    L_{lr} =  -(r(y^s) - r(\hat{y})) \sum_{t=1}^n \log p(y_t^s \mid y_1^s, \dots, y_{t-1}^s, x)
\end{equation*}
where $\hat{y}$ is the baseline output and $r$ is the reward function (i.e. the ROUGE metric). 

To increase the evaluation scores of generated summaries without hurting their human-readability or the relevance. The authors propose a mixed learning objective function that combines two objectives:
\begin{equation*}
    L_{mixed} = \gamma L_{rl} + (1 - \gamma) L_{ml}
\end{equation*}

The authors perform human evaluation on the relevance (how well does the summary capture the important parts of the article) and the readability (how well-written the summary is) of output summary. 
Five human evaluators were asked to allocate 1 to 10 for each abstract generated.
On both metrics, the mixed objective achieves the highest average scores 7.04 (readability) and 7.45 (relevance). For the RL objective, the average scores are 4.18 and 6.32. For the teacher forcing objective, the scores are 6.76 and 7.14.

\subsubsection[Summarize From Human Feedback]{Summarize From Human Feedback \citep{stiennon2020learning}}
\cite{stiennon2020learning} point out that both the human reference and the ROUGE metric are rough proxies for summary quality, which is what we really care about. Based on this idea, the authors propose to improve summary quality by training a summarization model to optimize for human preferences directly.
This is very different from previous training approaches where summarization models are trained to mimic human demonstrations (i.e. the reference summaries).
There are three steps in \cite{stiennon2020learning}'s approach:
\begin{enumerate}
    \item \textit{Collect Human Feedback}: First, sample a post $x$ from the Reddit TL;DR dataset \citep{volske-etal-2017-tl}, and use various policies to generate summaries of the selected Reddit posts. Then, select two summaries for human evaluation. The human evaluator will decide which summary is better.
    \item \textit{Train Reward Model}: The reward model $r_{\theta}$ takes a Reddit post $x$ and summary $y$ as input and output a scalar reward, which correlates with the input summary quality. In order to train such reward model, the loss function is defined as follows:
    \begin{equation*}
        \text{loss}(r_{\theta}) = \mathbb{E}_{(x,y_0,y_1,i)\sim D}[\log{(\sigma{( r_{\theta}(x,y_i) - r_{\theta}(x,y_{1-i}) )})}]
    \end{equation*}
    where $D$ is the dataset of human judgments. $i \in \{0, 1\}$ is the human preference label.
    \item \textit{Train Policy with PPO}: Given the reward model obtained last step, a policy is trained to generate high quality summaries that are in line with human preferences. The policy is trained using the PPO algorithm \cite{schulman2017proximal}. Besides the output from the reward model, the authors also include a KL divergence term in the final reward:
    \begin{equation*}
        R(x,y) = r_{\theta}(x,y) - \beta \log{[\pi_{}^{\text{RL}}(y\mid x) / \pi^{\text{SFT}}(y\mid x)]}
    \end{equation*}
    where $\pi^{\text{RL}}$ is the learned policy and $\pi^{\text{SFT}}$ is the supervised model. This term ensures the learned policy does not deviate too much from references.
\end{enumerate}

\paragraph{Results} On the TL;DR dataset, the proposed model achieves 61\% preference score (from human judges) against reference summaries. It significantly outperforms the supervised baseline's score 43\%.

\section{Factual Consistency of Abstractive Summarization Systems}
Despite significant improvements over previous methods in terms of automatic evaluation scores such as ROUGE \citep{lin-2004-rouge}, it is still challenging to ensure factual consistency of the generated summary with respect to the source. 
Table \ref{table:samples} shows an example of an output summary contains factual errors. 
One reason for this problem is that these $n$-gram overlapping metrics do not guarantee semantic correctness of generated summaries. For example, \citet{cao2018faithful} claims that about 30\% of summaries generated by abstractive models contain factual errors. \citet{maynez-etal-2020-faithfulness} discovered that 64.1\% of the summaries generated by a BERT-based abstractive summarization model on \textsc{XSum} contain hallucinations. Therefore, it is critical to detect factual errors introduced by abstractive summarization systems.

Unfortunately, determining factual correctness of summaries is incredibly difficult even for human beings. 
In this section, we will discuss some recently proposed factuality evaluation methods for abstractive summarization systems.

\definecolor{MyBlue}{rgb}{0.18,0.33,0.80}

\begin{table*}[t]
\small
\renewcommand{\arraystretch}{1.1}
\centering
\begin{tabular}{p{15.5cm}}
  \toprule
  \emph{Article}: Jerusalem (CNN)The flame of remembrance burns in Jerusalem, and a song of memory haunts Valerie Braham as it never has before. (...) ``Now I truly understand everyone who has lost a loved one,'' Braham said. \textcolor{MyBlue}{Her husband, Philippe Braham, was one of 17 people killed in January's terror attacks in Paris.} He was in a kosher supermarket when a gunman stormed in, killing four people, all of them Jewish. (...) \\ 
  \emph{Summary}: {\bf Valerie} braham was one of 17 people killed in January's terror attacks in Paris. \\
  \bottomrule
\end{tabular}
\caption{\label{table:samples} An example of a model output summary contains factual errors. In this example, it should be Philip Brabham, not Valerie Brabham, who was killed in the terrorist attack.}
\end{table*}

\subsection{Factual Consistency vs. Correctness}
It is important to distinguish between factual consistency and correctness. The two are not necessarily equal.
The factual consistency of the summary is determined by its consistency with the facts in the source document. It focuses on whether the summary presents the content of the article truthfully and accurately.
In contrast, factual correctness focuses on consistency with the facts in some external knowledge base (i.e. world knowledge). For instance, news articles with incorrect facts can be summarized with perfect factual consistency.

\subsection{Recent Approaches}
\subsubsection[Fact-conditioned Generation]{Fact-conditioned Generation \citep{cao2018faithful}}
\cite{cao2018faithful} propose to improve the factual consistency of summarization models by conditioning on facts extracted from the source document through the ``dual-attention'' mechanism.

\paragraph{Fact Description Extraction} The first step of \cite{cao2018faithful}'s approach is to extract fact descriptions from the source document. They use Open Information Extraction (OpenIE) to extract relation triples (subject; predicate; object) in the source document and concatenate them together as text descriptions. Then, they use a dependency parser to extract the the (subject; predicate) or (predicate; object) tuples that are not captured by OpenIE. 

\paragraph{Fact Aware Summarization Model} \cite{cao2018faithful}'s model consists of two GRU-based encoders: sentence encoder and relation encoder. The sentence encoder encodes the input document and the relation encoder encodes the extracted fact descriptions. For the decoder, since there are both document and relation representations as input, they develop two attentional layers to construct the overall context vector. This is referred as the ``dual attention'' mechanism.

\paragraph{Results} The authors perform human evaluation of factual consistency on 100 samples from the Gigaword test set, and find that the percentage of factual consistent summaries (judged by human) improves from 68\% to 87\%.

\subsubsection[QA-based Evaluation Methods]{QA-based Evaluation Methods \citep{durmus-etal-2020-feqa, wang-etal-2020-asking}}
Question Answering (QA) based evaluation methods have been shown effective in evaluating the factual consistency of generated summaries. The idea behind these methods is that if the summary and the input document are consistent with respect to an event, then when a question about the event is given, they should produce the same answer. Therefore, the more the summary and the source document produce the same answers, the more consistent they are. There are three important components in these methods: \textit{Question Generation Module}, \textit{Question Answering Module} and \textit{Answer Similarity Function}.
\paragraph{Question Generation Module} In order to automatically generate questions given the summary, \cite{durmus-etal-2020-feqa} mask all the noun phrases and named entities in the summary sentence. Each masked text span is considered as the gold standard answer. To generate the questions, \cite{durmus-etal-2020-feqa} fine-tune a BART language model on the QA2D dataset \citep{demszky2018transforming}. \cite{wang-etal-2020-asking} follows very similar approach for question generation. They fine-tune a BART language model on the NewsQA dataset,  a dataset consisting of CNN articles and human-written questions.

\paragraph{Question Answering Module} Given a question and answer pair based on a summary, \cite{durmus-etal-2020-feqa} and \cite{wang-etal-2020-asking} use off-the-shelf QA model to generate answers to the questions from the source document.

\paragraph{Answer Similarity Function} Both \cite{durmus-etal-2020-feqa} and \cite{wang-etal-2020-asking} use token-level F1 score to evaluate the generated answers. The final faithfulness score is given by averaging the answer similarity metric over all generated questions.

On human-annotated summary faithfulness datasets, both their methods achieve higher correlation scores compared with the ROUGE metric.

\subsubsection[FactCC]{FactCC \citep{kryscinski-etal-2020-evaluating}}
\cite{kryscinski-etal-2020-evaluating} propose a weakly-supervised approach for verifying factual consistency between input documents and generated summaries. Their basic idea is to train a natural language inference model on artificially created data. The synthetic training data generation process can be summarized as follows:
\begin{enumerate}
   \item Define semantically invariant and variant text transformations. The authors propose sentence negation, entity, pronoun, and number swaps as the variant transformations. 
   For entity and number swap, the entity in the claim sentence is replaced with a different entity in the document. For pronoun swap, the selected pronoun is swapped with a different one from the same pronoun group.
   
   To perform semantically invariant transformation, they first translate the sentence into another language (e.g. French, German, Chinese), and then translate it back to English.
    \item Given an input document $d$, randomly sample a sentence $s$ from $d$. Add ($d$, $s$, $+$) to the dataset. Here $+$ means the document and the input sentence are consistent.
    \item For each text transformation, apply it to the claim sentence $s$ and get a new sentence $s_{\text{new}}$. If the selected text transformation is semantically invariant, add ($d$, $s$, $+$) to the dataset. Otherwise, add ($d$, $s$, $-$).
\end{enumerate}

\paragraph{Results} After the data creation process, a factual consistency classifier is trained on the synthetic data. The authors labeled 931 examples as the validation set and 503 examples as the test set.
On the test set, their best model achieves 72.88\% weighted accuracy, which is about 20\% higher than the BERT-based NLI baseline model.
One limitation of \cite{kryscinski-etal-2020-evaluating}'s approach is that it cannot correctly classify examples where the generated summary is highly abstractive.




\section{Discussion}
Despite significant efforts made by the research community, there are still many challenges in abstractive summarization: 1) it is challenging to ensure factual consistency of the generated summary with respect to the source. There are two main reasons for this:

First, it is difficult to design fast and low-cost factuality-aware evaluation metrics for summarization models. Mostly used $n$-gram evaluation approach does not take factuality consistency into consideration. On the other hand, the human evaluation process is too expensive and time-consuming. Second, many abstractive summarization datasets are automatically generated, which inevitably contain noise. Neural network models can easily overfit these noises during training and generate non-factual errors during the inference process.
3) As the scale of the model grows, so does the demand for training data. It requires researchers to find sample-efficient training methods. One possible direction is to find pre-training objectives with good inductive bias towards summarization tasks.

\section{Conclusion}
In this survey, we have summarized and analyzed state-of-the-art neural network-based abstractive summarization methods. We also point out that there are many challenges of these models remain unsolved, especially the issue of factual consistency between the summary and the source. We hope that this survey can help researchers better understand the development of abstractive summarization in recent years, so as to make more scientifically meaningful progress in this field.


\bibliographystyle{plainnat}
\bibliography{references}

\begin{thebibliography}{35}
\providecommand{\natexlab}[1]{#1}
\providecommand{\url}[1]{\texttt{#1}}
\expandafter\ifx\csname urlstyle\endcsname\relax
  \providecommand{\doi}[1]{doi: #1}\else
  \providecommand{\doi}{doi: \begingroup \urlstyle{rm}\Url}\fi

\bibitem[Bahdanau et~al.(2015)Bahdanau, Cho, and
  Bengio]{DBLP:journals/corr/BahdanauCB14}
Dzmitry Bahdanau, Kyunghyun Cho, and Yoshua Bengio.
\newblock Neural machine translation by jointly learning to align and
  translate.
\newblock In Yoshua Bengio and Yann LeCun, editors, \emph{3rd International
  Conference on Learning Representations, {ICLR} 2015, San Diego, CA, USA, May
  7-9, 2015, Conference Track Proceedings}, 2015.
\newblock URL \url{http://arxiv.org/abs/1409.0473}.

\bibitem[Banko et~al.(2000)Banko, Mittal, and
  Witbrock]{10.3115/1075218.1075259}
Michele Banko, Vibhu~O. Mittal, and Michael~J. Witbrock.
\newblock Headline generation based on statistical translation.
\newblock In \emph{Proceedings of the 38th Annual Meeting on Association for
  Computational Linguistics}, ACL '00, page 318–325, USA, 2000. Association
  for Computational Linguistics.
\newblock \doi{10.3115/1075218.1075259}.
\newblock URL \url{https://doi.org/10.3115/1075218.1075259}.

\bibitem[Cao et~al.(2018)Cao, Wei, Li, and Li]{cao2018faithful}
Ziqiang Cao, Furu Wei, Wenjie Li, and Sujian Li.
\newblock Faithful to the original: Fact aware neural abstractive
  summarization.
\newblock In \emph{Thirty-Second AAAI Conference on Artificial Intelligence},
  2018.

\bibitem[Chopra et~al.(2016)Chopra, Auli, and
  Rush]{chopra-etal-2016-abstractive}
Sumit Chopra, Michael Auli, and Alexander~M. Rush.
\newblock Abstractive sentence summarization with attentive recurrent neural
  networks.
\newblock In \emph{Proceedings of the 2016 Conference of the North {A}merican
  Chapter of the Association for Computational Linguistics: Human Language
  Technologies}, pages 93--98, San Diego, California, June 2016. Association
  for Computational Linguistics.
\newblock \doi{10.18653/v1/N16-1012}.
\newblock URL \url{https://aclanthology.org/N16-1012}.

\bibitem[Chung et~al.(2014)Chung, Gulcehre, Cho, and
  Bengio]{69e088c8129341ac89810907fe6b1bfe}
Junyoung Chung, Caglar Gulcehre, Kyunghyun Cho, and Yoshua Bengio.
\newblock Empirical evaluation of gated recurrent neural networks on sequence
  modeling.
\newblock In \emph{NIPS 2014 Workshop on Deep Learning, December 2014}, 2014.

\bibitem[Demszky et~al.(2018)Demszky, Guu, and Liang]{demszky2018transforming}
Dorottya Demszky, Kelvin Guu, and Percy Liang.
\newblock Transforming question answering datasets into natural language
  inference datasets.
\newblock \emph{arXiv preprint arXiv:1809.02922}, 2018.

\bibitem[Devlin et~al.(2019)Devlin, Chang, Lee, and
  Toutanova]{devlin-etal-2019-bert}
Jacob Devlin, Ming-Wei Chang, Kenton Lee, and Kristina Toutanova.
\newblock {BERT}: Pre-training of deep bidirectional transformers for language
  understanding.
\newblock In \emph{Proceedings of the 2019 Conference of the North {A}merican
  Chapter of the Association for Computational Linguistics: Human Language
  Technologies, Volume 1 (Long and Short Papers)}, pages 4171--4186,
  Minneapolis, Minnesota, June 2019. Association for Computational Linguistics.
\newblock \doi{10.18653/v1/N19-1423}.
\newblock URL \url{https://aclanthology.org/N19-1423}.

\bibitem[Durmus et~al.(2020)Durmus, He, and Diab]{durmus-etal-2020-feqa}
Esin Durmus, He~He, and Mona Diab.
\newblock {FEQA}: A question answering evaluation framework for faithfulness
  assessment in abstractive summarization.
\newblock In \emph{Proceedings of the 58th Annual Meeting of the Association
  for Computational Linguistics}, pages 5055--5070, Online, July 2020.
  Association for Computational Linguistics.
\newblock \doi{10.18653/v1/2020.acl-main.454}.
\newblock URL \url{https://aclanthology.org/2020.acl-main.454}.

\bibitem[Elman(1990)]{elman1990finding}
Jeffrey~L Elman.
\newblock Finding structure in time.
\newblock \emph{Cognitive science}, 14\penalty0 (2):\penalty0 179--211, 1990.

\bibitem[Gehrmann et~al.(2018)Gehrmann, Deng, and
  Rush]{gehrmann-etal-2018-bottom}
Sebastian Gehrmann, Yuntian Deng, and Alexander Rush.
\newblock Bottom-up abstractive summarization.
\newblock In \emph{Proceedings of the 2018 Conference on Empirical Methods in
  Natural Language Processing}, pages 4098--4109, Brussels, Belgium,
  October-November 2018. Association for Computational Linguistics.
\newblock \doi{10.18653/v1/D18-1443}.
\newblock URL \url{https://aclanthology.org/D18-1443}.

\bibitem[Graff et~al.(2003)Graff, Kong, Chen, and Maeda]{graff2003english}
David Graff, Junbo Kong, Ke~Chen, and Kazuaki Maeda.
\newblock English gigaword.
\newblock \emph{Linguistic Data Consortium, Philadelphia}, 4\penalty0
  (1):\penalty0 34, 2003.

\bibitem[Hochreiter and Schmidhuber(1997)]{HochSchm97}
Sepp Hochreiter and Jürgen Schmidhuber.
\newblock Long short-term memory.
\newblock \emph{Neural Computation}, 9\penalty0 (8):\penalty0 1735--1780, 1997.

\bibitem[Jean et~al.(2015)Jean, Cho, Memisevic, and
  Bengio]{jean-etal-2015-using}
S{\'e}bastien Jean, Kyunghyun Cho, Roland Memisevic, and Yoshua Bengio.
\newblock On using very large target vocabulary for neural machine translation.
\newblock In \emph{Proceedings of the 53rd Annual Meeting of the Association
  for Computational Linguistics and the 7th International Joint Conference on
  Natural Language Processing (Volume 1: Long Papers)}, pages 1--10, Beijing,
  China, July 2015. Association for Computational Linguistics.
\newblock \doi{10.3115/v1/P15-1001}.
\newblock URL \url{https://aclanthology.org/P15-1001}.

\bibitem[Knight and Marcu(2002)]{knight2002summarization}
Kevin Knight and Daniel Marcu.
\newblock Summarization beyond sentence extraction: A probabilistic approach to
  sentence compression.
\newblock \emph{Artificial Intelligence}, 139\penalty0 (1):\penalty0 91--107,
  2002.

\bibitem[Kryscinski et~al.(2020)Kryscinski, McCann, Xiong, and
  Socher]{kryscinski-etal-2020-evaluating}
Wojciech Kryscinski, Bryan McCann, Caiming Xiong, and Richard Socher.
\newblock Evaluating the factual consistency of abstractive text summarization.
\newblock In \emph{Proceedings of the 2020 Conference on Empirical Methods in
  Natural Language Processing (EMNLP)}, pages 9332--9346, Online, November
  2020. Association for Computational Linguistics.
\newblock \doi{10.18653/v1/2020.emnlp-main.750}.
\newblock URL \url{https://aclanthology.org/2020.emnlp-main.750}.

\bibitem[Lewis et~al.(2020)Lewis, Liu, Goyal, Ghazvininejad, Mohamed, Levy,
  Stoyanov, and Zettlemoyer]{lewis-etal-2020-bart}
Mike Lewis, Yinhan Liu, Naman Goyal, Marjan Ghazvininejad, Abdelrahman Mohamed,
  Omer Levy, Veselin Stoyanov, and Luke Zettlemoyer.
\newblock {BART}: Denoising sequence-to-sequence pre-training for natural
  language generation, translation, and comprehension.
\newblock In \emph{Proceedings of the 58th Annual Meeting of the Association
  for Computational Linguistics}, pages 7871--7880, Online, July 2020.
  Association for Computational Linguistics.
\newblock \doi{10.18653/v1/2020.acl-main.703}.
\newblock URL \url{https://aclanthology.org/2020.acl-main.703}.

\bibitem[Lin(2004)]{lin-2004-rouge}
Chin-Yew Lin.
\newblock {ROUGE}: A package for automatic evaluation of summaries.
\newblock In \emph{Text Summarization Branches Out}, pages 74--81, Barcelona,
  Spain, July 2004. Association for Computational Linguistics.
\newblock URL \url{https://aclanthology.org/W04-1013}.

\bibitem[Liu and Lapata(2019)]{liu-lapata-2019-text}
Yang Liu and Mirella Lapata.
\newblock Text summarization with pretrained encoders.
\newblock In \emph{Proceedings of the 2019 Conference on Empirical Methods in
  Natural Language Processing and the 9th International Joint Conference on
  Natural Language Processing (EMNLP-IJCNLP)}, pages 3730--3740, Hong Kong,
  China, November 2019. Association for Computational Linguistics.
\newblock \doi{10.18653/v1/D19-1387}.
\newblock URL \url{https://aclanthology.org/D19-1387}.

\bibitem[Maynez et~al.(2020)Maynez, Narayan, Bohnet, and
  McDonald]{maynez-etal-2020-faithfulness}
Joshua Maynez, Shashi Narayan, Bernd Bohnet, and Ryan McDonald.
\newblock On faithfulness and factuality in abstractive summarization.
\newblock In \emph{Proceedings of the 58th Annual Meeting of the Association
  for Computational Linguistics}, pages 1906--1919, Online, July 2020.
  Association for Computational Linguistics.
\newblock \doi{10.18653/v1/2020.acl-main.173}.
\newblock URL \url{https://www.aclweb.org/anthology/2020.acl-main.173}.

\bibitem[Nallapati et~al.(2016)Nallapati, Zhou, dos Santos, Gulcehre, and
  Xiang]{nallapati-etal-2016-abstractive}
Ramesh Nallapati, Bowen Zhou, Cicero dos Santos, Caglar Gulcehre, and Bing
  Xiang.
\newblock Abstractive text summarization using sequence-to-sequence {RNN}s and
  beyond.
\newblock In \emph{Proceedings of The 20th {SIGNLL} Conference on Computational
  Natural Language Learning}, pages 280--290, Berlin, Germany, August 2016.
  Association for Computational Linguistics.
\newblock \doi{10.18653/v1/K16-1028}.
\newblock URL \url{https://aclanthology.org/K16-1028}.

\bibitem[Napoles et~al.(2012)Napoles, Gormley, and
  Van~Durme]{napoles-etal-2012-annotated}
Courtney Napoles, Matthew Gormley, and Benjamin Van~Durme.
\newblock Annotated {G}igaword.
\newblock In \emph{Proceedings of the Joint Workshop on Automatic Knowledge
  Base Construction and Web-scale Knowledge Extraction ({AKBC}-{WEKEX})}, pages
  95--100, Montr{\'e}al, Canada, June 2012. Association for Computational
  Linguistics.
\newblock URL \url{https://aclanthology.org/W12-3018}.

\bibitem[Narayan et~al.(2018)Narayan, Cohen, and
  Lapata]{narayan-etal-2018-dont}
Shashi Narayan, Shay~B. Cohen, and Mirella Lapata.
\newblock Don{'}t give me the details, just the summary! topic-aware
  convolutional neural networks for extreme summarization.
\newblock In \emph{Proceedings of the 2018 Conference on Empirical Methods in
  Natural Language Processing}, pages 1797--1807, Brussels, Belgium,
  October-November 2018. Association for Computational Linguistics.
\newblock \doi{10.18653/v1/D18-1206}.
\newblock URL \url{https://www.aclweb.org/anthology/D18-1206}.

\bibitem[Nenkova and McKeown(2011)]{nenkova2011automatic}
Ani Nenkova and Kathleen McKeown.
\newblock \emph{Automatic summarization}.
\newblock Now Publishers Inc, 2011.

\bibitem[Paulus et~al.(2018)Paulus, Xiong, and Socher]{paulus2018a}
Romain Paulus, Caiming Xiong, and Richard Socher.
\newblock A deep reinforced model for abstractive summarization.
\newblock In \emph{International Conference on Learning Representations}, 2018.
\newblock URL \url{https://openreview.net/forum?id=HkAClQgA-}.

\bibitem[Ranzato et~al.(2016)Ranzato, Chopra, Auli, and
  Zaremba]{DBLP:journals/corr/RanzatoCAZ15}
Marc'Aurelio Ranzato, Sumit Chopra, Michael Auli, and Wojciech Zaremba.
\newblock Sequence level training with recurrent neural networks.
\newblock In Yoshua Bengio and Yann LeCun, editors, \emph{4th International
  Conference on Learning Representations, {ICLR} 2016, San Juan, Puerto Rico,
  May 2-4, 2016, Conference Track Proceedings}, 2016.
\newblock URL \url{http://arxiv.org/abs/1511.06732}.

\bibitem[Rennie et~al.(2017)Rennie, Marcheret, Mroueh, Ross, and
  Goel]{rennie2017self}
Steven~J Rennie, Etienne Marcheret, Youssef Mroueh, Jerret Ross, and Vaibhava
  Goel.
\newblock Self-critical sequence training for image captioning.
\newblock In \emph{Proceedings of the IEEE conference on computer vision and
  pattern recognition}, pages 7008--7024, 2017.

\bibitem[Rush et~al.(2015)Rush, Chopra, and Weston]{rush-etal-2015-neural}
Alexander~M. Rush, Sumit Chopra, and Jason Weston.
\newblock A neural attention model for abstractive sentence summarization.
\newblock In \emph{Proceedings of the 2015 Conference on Empirical Methods in
  Natural Language Processing}, pages 379--389, Lisbon, Portugal, September
  2015. Association for Computational Linguistics.
\newblock \doi{10.18653/v1/D15-1044}.
\newblock URL \url{https://aclanthology.org/D15-1044}.

\bibitem[Schulman et~al.(2017)Schulman, Wolski, Dhariwal, Radford, and
  Klimov]{schulman2017proximal}
John Schulman, Filip Wolski, Prafulla Dhariwal, Alec Radford, and Oleg Klimov.
\newblock Proximal policy optimization algorithms.
\newblock \emph{arXiv preprint arXiv:1707.06347}, 2017.

\bibitem[See et~al.(2017)See, Liu, and Manning]{see-etal-2017-get}
Abigail See, Peter~J. Liu, and Christopher~D. Manning.
\newblock Get to the point: Summarization with pointer-generator networks.
\newblock In \emph{Proceedings of the 55th Annual Meeting of the Association
  for Computational Linguistics (Volume 1: Long Papers)}, pages 1073--1083,
  Vancouver, Canada, July 2017. Association for Computational Linguistics.
\newblock \doi{10.18653/v1/P17-1099}.
\newblock URL \url{https://aclanthology.org/P17-1099}.

\bibitem[{Stiennon} et~al.(2020){Stiennon}, {Ouyang}, {Wu}, {Ziegler}, {Lowe},
  {Voss}, {Radford}, {Amodei}, and {Christiano}]{stiennon2020learning}
Nisan {Stiennon}, Long {Ouyang}, Jeffrey {Wu}, Daniel~M. {Ziegler}, Ryan
  {Lowe}, Chelsea {Voss}, Alec {Radford}, Dario {Amodei}, and Paul~F.
  {Christiano}.
\newblock Learning to summarize from human feedback.
\newblock In \emph{Advances in Neural Information Processing Systems},
  volume~33, pages 3008--3021, 2020.

\bibitem[Vaswani et~al.(2017)Vaswani, Shazeer, Parmar, Uszkoreit, Jones, Gomez,
  Kaiser, and Polosukhin]{NIPS2017_3f5ee243}
Ashish Vaswani, Noam Shazeer, Niki Parmar, Jakob Uszkoreit, Llion Jones,
  Aidan~N Gomez, \L~ukasz Kaiser, and Illia Polosukhin.
\newblock Attention is all you need.
\newblock In I.~Guyon, U.~V. Luxburg, S.~Bengio, H.~Wallach, R.~Fergus,
  S.~Vishwanathan, and R.~Garnett, editors, \emph{Advances in Neural
  Information Processing Systems}, volume~30. Curran Associates, Inc., 2017.
\newblock URL
  \url{https://proceedings.neurips.cc/paper/2017/file/3f5ee243547dee91fbd053c1c4a845aa-Paper.pdf}.

\bibitem[V{\"o}lske et~al.(2017)V{\"o}lske, Potthast, Syed, and
  Stein]{volske-etal-2017-tl}
Michael V{\"o}lske, Martin Potthast, Shahbaz Syed, and Benno Stein.
\newblock {TL};{DR}: Mining {R}eddit to learn automatic summarization.
\newblock In \emph{Proceedings of the Workshop on New Frontiers in
  Summarization}, pages 59--63, Copenhagen, Denmark, September 2017.
  Association for Computational Linguistics.
\newblock \doi{10.18653/v1/W17-4508}.
\newblock URL \url{https://aclanthology.org/W17-4508}.

\bibitem[Wang et~al.(2020)Wang, Cho, and Lewis]{wang-etal-2020-asking}
Alex Wang, Kyunghyun Cho, and Mike Lewis.
\newblock Asking and answering questions to evaluate the factual consistency of
  summaries.
\newblock In \emph{Proceedings of the 58th Annual Meeting of the Association
  for Computational Linguistics}, pages 5008--5020, Online, July 2020.
  Association for Computational Linguistics.
\newblock \doi{10.18653/v1/2020.acl-main.450}.
\newblock URL \url{https://aclanthology.org/2020.acl-main.450}.

\bibitem[Williams and Zipser(1989)]{6795228}
Ronald~J. Williams and David Zipser.
\newblock A learning algorithm for continually running fully recurrent neural
  networks.
\newblock \emph{Neural Computation}, 1\penalty0 (2):\penalty0 270--280, 1989.
\newblock \doi{10.1162/neco.1989.1.2.270}.

\bibitem[Zhang et~al.(2020)Zhang, Zhao, Saleh, and Liu]{pmlr-v119-zhang20ae}
Jingqing Zhang, Yao Zhao, Mohammad Saleh, and Peter Liu.
\newblock {PEGASUS}: Pre-training with extracted gap-sentences for abstractive
  summarization.
\newblock In Hal~Daumé III and Aarti Singh, editors, \emph{Proceedings of the
  37th International Conference on Machine Learning}, volume 119 of
  \emph{Proceedings of Machine Learning Research}, pages 11328--11339. PMLR,
  13--18 Jul 2020.
\newblock URL \url{https://proceedings.mlr.press/v119/zhang20ae.html}.

\end{thebibliography}

\end{document}